# Deep Odometry Systems on Edge with EKF-LoRa Backend for Real-Time Positioning in Adverse Environment

Zhuangzhuang Dai, Muhamad Risqi U. Saputra, Chris Xiaoxuan Lu, Andrew Markham, and Niki Trigoni

*Abstract*— Ubiquitous positioning for pedestrian in adverse environment has served a long standing challenge. Despite dramatic progress made by Deep Learning, multi-sensor deep odometry systems yet pose a high computational cost and suffer from cumulative drifting errors over time. Thanks to the increasing computational power of edge devices, we propose a novel ubiquitous positioning solution by integrating state-of-the-art deep odometry models on edge with an EKF (Extended Kalman Filter)-LoRa backend. We carefully compare and select three sensor modalities, i.e., an Inertial Measurement Unit (IMU), a millimetre-wave (mmWave) radar, and a thermal infrared camera, and realise their deep odometry inference engines which runs in real-time. A pipeline of deploying deep odometry considering accuracy, complexity, and edge platform is proposed. We design a LoRa link for positional data backhaul and projecting aggregated positions of deep odometry into the global frame. We find that a simple EKF based fusion module is sufficient for generic positioning calibration with over 34% accuracy gains against any standalone deep odometry system. Extensive tests in different environments validate the efficiency and efficacy of our proposed positioning system.

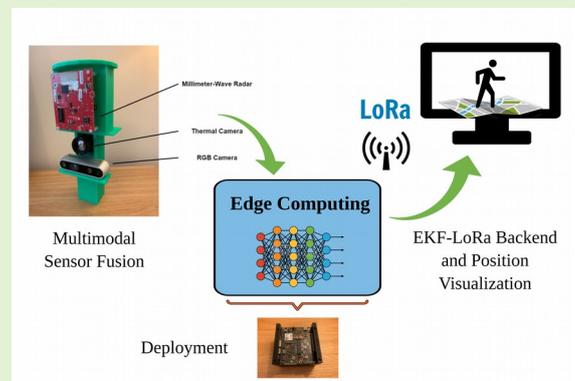

## I. INTRODUCTION

POSITIONING systems play a key role in human-centric technologies. However, prevailing positioning systems, such as satellite navigation and radio positioning, depend on elaborated infrastructure as a prerequisite. Fingerprint-based localization requires expensive site-survey yet is restricted to a specific area. Recently, more research interests focus on ubiquitous positioning where out-of-box positional information is generated regardless of the environment. Search and rescue operation, underground inspection, and subterranean environment exploration will all benefit significantly from such positioning systems with limited or no access to pre-installed access points. Whereby, the use of infrastructure-free solutions such as cameras and Micro-Electro-Mechanical Systems (MEMS) sensors, e.g., Inertial Measurement Units (IMU), have witnessed an outburst in the past decade.

The IMU is a particularly low-cost, low-power motion sensor which lends itself a ubiquitous candidate in modern active devices. Nevertheless, owing to its ego-centric nature the IMU incurs a large amount of noise and intrinsic biases. For instance, state-of-the-art Inertial Navigation System (INS) which only uses the IMU suffers from large drifting errors after a fairly short contiguous use. Visual cameras, on the other hand, create rich visual data from the surroundings. Visual Odometry (VO) based on either feature points or direct tracking has demonstrated strengths in ubiquitous positioning [38]. With the aid of Deep Learning and visual-inertial data fusion [3], a sub-meter precision has been achieved. However, VO faces challenges posed by visual degradation such as smoke, glare, and darkness. Even if when VO is combined with depth cameras (RGB-D) or LiDAR (Light Detection and Ranging) [39], it will still suffer in smoky environment. Emerging sensors in conjunction with complex Deep Neural Networks (DNN) are recently proposed to tackle these problems, namely thermal infrared camera [34] and mmWave radar [42]. Thermal infrared cameras are able to detect radiation of objects so as to gain vision even in darkness. On the other hand, mmWave radar generates sparse point clouds that can penetrate smokes.

This research has been financially supported by the National Institute of Standards and Technology (NIST) via the grant Pervasive, Accurate, and Reliable Location-based Services for Emergency Responders (Federal Grant: 70NANB17H185).

Z. Dai, A. Markham, and N. Trigoni are with University at Oxford, Oxford, England, United Kingdom, e-mail: firstname.surname@cs.ox.ac.uk.

M. R. U. Saputra is with Monash University Indonesia, Tangerang, Indonesia, e-mail: risqi.saputra@monash.edu.

Chris Xiaoxuan Lu is with University of Edinburgh, Edinburgh, Scotland, United Kingdom, e-mail: xiaoxuan.lu@ed.ac.uk.

Those sensor modalities can be an alternative solution for positioning in adverse environment.

Despite the capabilities to infer position in adverse environment, positioning module equipped with thermal or mmWave radar must process the sensor data at the edge device, especially when cloud or server off-loading is not possible [18]. Though the aforementioned deep learning based multi-modality odometry systems prove capable of tracking humans or robots accurately, running such complex DNNs on resource-constrained devices in real-time remains non-trivial. Few studies have investigated the real-time performance of applying these deep odometry systems [19]. To be specific, on-device computing is often confronted by a lack of processing power, a lack of memory space, and a high demand for power. To the best of our knowledge, never has any researches worked out the best practices of deploying these deep odometry under different resource constraints.

Besides looking at the complexity of deploying deep odometry to resource constrained devices, odometry systems often pose with drifting problem and they aggregate positional changes based on local coordinates. To project a deep odometry positioning module to the global coordinate system, a marker or an absolute initial position is usually used to bridge the gap [20]. Since we want to build a ubiquitous positioning system for out-of-box usage, GNSS or preset markers are excluded. Existing wireless techniques, such as Wi-Fi, LTE [1], and Ultra-Wide Band (UWB) [26], are widely used for range and radio either require pre-installed base stations or have limited range. We determine LoRa makes an ideal choice of wireless communication for: LoRa has an excellent range in NLOS environments with a sufficient and highly-configurable bandwidth for backhaul [4]; A LoRa access point (AP) can be a gateway or simply another LoRa node, either of which is low-power and easy to setup. Based on a LoRa link, Time of Flight (TOF), Angle-of-Arrival (AOA), as well as Power Delay Profile (PDP) can be extracted at the receiver to derive accurate positions. However, these require elaborate synchronisation and costly hardware setup [2]. We advocate a device-free metric, Received Signal Strength Indicator (RSSI), is everything needed for positioning calibration [32]. An RSSI reading comes with every received LoRa packet which can be used to reveal the range. Since current deep odometry systems already accomplish great accuracy, we propose an EKF-based backend to fuse the odometry trajectory with LoRa RSSIs for a robust ubiquitous positioning system. The basic hypothesis is that even with only one LoRa receiver, a coarse range estimate will be able to calibrate the odometry in the long run.

In this work, we put forward a novel ubiquitous positioning solution, which integrates deep odometry on edge, LoRa, and an augmented EKF, to pedestrian localization in unsurveyed and visually-degraded environments. We first study the state-of-the-art deep odometry systems in terms of accuracy, computational complexity, robustness in adverse environment, and frame rate. We evaluate their performance on edge platforms with various resource constraints, and propose a universal pipeline to deployment them. We believe this is significant for bringing deep odometry systems to real-world positioning applications. Thereafter, we propose a LoRa-EKF backend for positional data backhaul, as well as drifting error calibration. Using EKF as backend is not only efficient but effective in containing the drifting errors of deep odometry in the long run. Our major contributions incorporate: (1) We evaluate embedded multi-modality deep odometry models for real-time positioning and propose best practices for deployment under different resource constraints; (2) We propose an EKF-LoRa backend for wireless backhaul and positioning calibration projected in the global frame[1]; (3) We validate the performance of the proposed solution on various edge devices in different environments.

The rest of this paper is organized as follows. Section II reviews state-of-the-art positioning techniques and narrates the motivations of our approach. The system integration is explained in details in Section III. Real-time evaluation results are presented in Section IV. Limitations and prospects of future work are discussed in Section V. Section VI summarizes this work.

## II. RELATED WORK

*1) Pedestrian Localization:* Pedestrian localization in challenging indoor environments has long served research attention. Systems based on IMU alone fail to realize long-term stability [11]. For instance, INS and PDR (Pedestrian Dead Reckoning) are based on physics-driven modelling of target motion [23]. It is inevitable drifting errors accumulate as time elapses [22]. On the other hand, visual cameras for navigation draw increasing interest for their wide use and low cost [11]. Nonetheless, state-of-the-art VO become ineffective dealing with featureless scenes or visual degradation [20]. In order to locate pedestrians in such challenging environment, Deep Learning based methods using emerging sensors become a hotspot recently.

Deep learning, i.e., a neural network with over three layers [36], draws increasing attention in the past decade for its excellence in handling sensor data. Monocular VO, such as ORB-SLAM [31], has seen groundbreaking progress in the past few years. C. Chen *et al.* [5] proposed IONet, a learning-based odometry that utilizes merely one IMU together with published dataset [6]. M. R. U. Saputra *et al.* [34] proposed DeepTIO, a hybrid deep learning network combining inertial, visual, and thermal odometry. C. X. Lu. *et al.* [29] proposed milliEgo, an odometry system based on feature fusion of a mmWave radar and an IMU. LiDAR generates dense point clouds which have been exploited for localization and mapping [39].

*2) Edge Computing and Model Compression:* Although learning-based odometry systems witness significant improvement in accuracy, they typically incur much higher computational costs which prevent an efficient real-time solution. There are over 1.3 million parameters in the base IONet [6]. The milliEgo [29] has over 33 million parameters; DeepTIO [34] possesses approximately 120 million parameters. They all take up an enormous amount of RAM (Random Access Memory) when processing. Cloud offloading is commonly used in dealing with such DNNs. However, access to cloud may not be available in emergency or post-disaster scenarios. Data uplink also requires an enormous bandwidth to upload raw sensory data which can be infeasible. The network latency can be hundreds of milliseconds, up to a few seconds [24], which impairs the instantaneity of the positioning system.

---
[1]The source code of this work has been made open-source on Github, availabe at: https://github.com/zdai257/LoRaRX .

The core idea of edge computing lies in performing calculations near the sensor or actuators' end. There is also need to carry out on-device data processing, storage, and visualisation for positioning services. In doing so, a key step is to optimise the DNN engines running on the edge devices. Knowledge distillation, proposed by G. Hinton *et al.* [21], aims to miniature DNN design. It transfers the knowledge of a larger teacher model to a smaller student model [33], in which student models are found sometimes outperform the teacher. Network Pruning tends to remove less significant parameters, e.g., weights close to 0, in the DNN. X. Ma *et al.* [30] proposed a sparsely pruning pattern that overcomes the weaknesses of non-structured, fine-grained pruning or structured pruning. Quantization [7] refers to truncating the DNN parameters, e.g., casting 32-bit floating-point weights to 8-bit integers. This effectively avoids costly floating-point multiplications so as to boosting the inference speed at the expense of a compromised accuracy.

Although trimming the network in the aforementioned ways may succeed in a one-off attempt [33], [37], it cannot be popularised to suit different models on different devices. Indeed, popular machine learning frameworks provide model compression toolbox as universal solutions of DNN optimization. TensorRT is an NVIDIA proprietary framework that enables optimized runtime performance in terms of computational latency and loading time. TensorRT utilises graph optimizations, layer fusion, and finding an efficient implementation of a DNN model to convert and run DNNs on mobile, embedded, and IoT devices. Nevertheless, TensorRT has limited support across different Machine Learning frameworks yet. In pursuit of better interoperability, ONNX [28] emerged as an open format established to represent Machine Learning models. This sheds light on converting deep odometry models developed in different frameworks to a universal representation.

*3) Wireless Comms and Backend:* Positioning data can be sent back to server through cellular or Wi-Fi network. In the case of emergency or post-disaster scenarios, these wireless networks are no longer reliable. Other contingent wireless communication links, such as Bluetooth [10] and UWB [26], suffer from limited ranges especially when transmitting through walls. LoRa is a Low-Power Wide-Area Network (LPWAN) physical layer operating in the license-free frequency bands [13]. Thanks to the Chirp Spread Spectrum (CSS) modulation scheme, LoRa achieves a multi-kilometer communication range and an ultra low power consumption even in built-up areas [16]. The CSS enables great robustness against noise and interference [4]. Moreover, configuration of the LoRa Spreading Factor (SF) [25] allows trading off coverage for data rate which ranges from $300bps$ to $27Kbps$. Despite a limited data rate, positioning data, e.g., 6-DoF (Six Degree of Freedom) poses, demand a low bandwidth in comparison to raw sensory data.

Additionally, the LoRa RSSI can be obtained without setup of any equipment. Various generic positioning fusion backends can be used for positioning fusion. Kalman Filter is lightweight, whereas, only effects in linear systems. In contrast, Particle Filter [40], Unscented Kalman Filter, and Sequential Monte Carlo Kalman Filter [27] deal with non-linear systems but impose high computational costs. EKF [9] is among the most widely used methods applicable to non-linear systems. Y. Zhuang *et al.* [43] built a generic indoor localisation system by fusing MEMS and wireless sensing of BLE (Bluetooth Low Energy), Wi-Fi, and other sources through EKF. S. M. M. Dehghan *et al.* [12] utilized EKF to fuse RSSIs measured by UAVs (Unmanned Aerial Vehicles) to locate unknown RF sources. We take inspirations from the above to integrate LoRa and EKF into a simple wireless position-calibrating backend.

## III. DEEP ODOMETRY ON EDGE FOR UBIQUITOUS POSITIONING

Deep learning-based odometry systems have surpassed model-based algorithms for ubiquitous positioning in adverse environment in all aspects, except their high computational cost which draws concern for their feasibility in real-time. In this work, we encapsulate state-of-the-art deep odometry models using emerging sensors that works well in adverse conditions, namely IMU only, mmWave radar + IMU, and thermal camera + IMU. We realize these DNNs on edge computing platforms with different resource constraints to validate their real-time performance. In doing so, we design a pipeline of deploying these models with respect to accuracy, reliability, and complexity. We develop a generic EKF-LoRa backend for positional data backhaul and projecting the egocentric positioning of odometry in the global frame. We determine an augmented EKF fusion module, together with the LoRa RSSI which requires no additional equipment, are sufficient to allow automatic positioning calibration.

### A. Sensors and Deep Odometry

Ubiquitous positioning in adverse environment requires sensors that persist in swift body motions, darkness, glare, smoke, and extreme temperatures. This rules out the prevailing VO, or satellite and terrestrial navigation systems which rely on existing infrastructure. On the other hand, ideal sensors should be relatively low-power and easy to be integrated in wearable devices. We shortlist emerging multi-sensor aided deep odometry candidates with proved capabilities in adverse environment, including IMU only, mmWave radar + IMU, and thermal camera + IMU. We contrast merits and drawbacks of the state-of-the-art models as shown in Table I.

As a first step, we 3D print a handheld device to hold these sensors as shown in Fig. 1. It coaxially aligns a TI AWR1843 mmWave radar, a Flir Boson 640 thermal camera, and an Xsens MTi 1-series IMU (together with an RGB-D camera for reference purpose). A Velodyne's Puck LiDAR is used for ground-truth data collection at $10Hz$.

The IMUs are pervasive in modern active devices to provide ego-motion sensing as they are compact, low-cost, and power efficient. The Xsens MTi 1-series IMU combines an accelerometer and a gyroscope with a sampling rate of $400Hz$. The IONet [5] is a particularly light-weight learning-based DNN to reconstruct 6-DoF translation and rotations, $\tilde{\mathbf{y}} = \tilde{\mathbf{t}}, \tilde{\mathbf{r}} \tilde{\mathbf{y}} \in \mathbb{R}$, from raw IMU data. The IONet is not only robust but extremely compact.

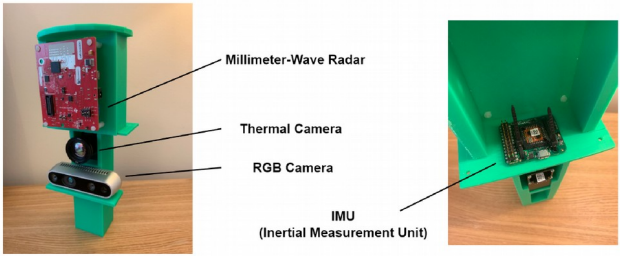

Fig. 1: A 3D printed handheld device (could be mounted on a helmet) to hold a mmWave radar, a thermal camera, an RGB-D camera, and an IMU.

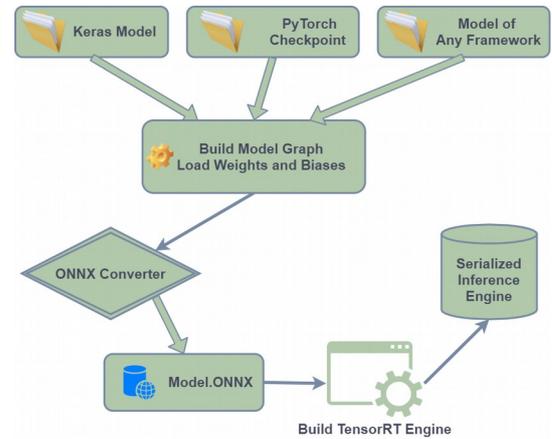

Fig. 2: A workflow of the deployment pipeline that converts deep odometry models to inference engines via ONNX.

The mmWave radar uses Frequency Modulated Continuous Wave (FMCW) in which a linear chirp signal allows distance estimation from reflectors. State-of-the-art milliEgo [29] exhibits excellent capability in estimating human ego-motion, immune to visual degradation. The mmWave radar generates sparse point clouds which takes up a low bandwidth compared to visual cameras. This allows a lightweight network design. On the downside, milliEgo requires reflectors within $5-10m$ to function and suffers from a limited range resolution. The angular resolution is also coarse using the single-chip AWR1843 mmWave radar. As a result, the generated point clouds are sparse and noisy.

Thermal infrared cameras capture the radiation emitted from objects in the Long-Wave Infrared (LWIR) portion of the spectrum [17]. The obtained raw radiometric data are then converted to a temperature profile represented in grayscale in the Flir Boson thermal camera. The strength of thermal camera lies in its immunity to illumination conditions allowing perceive object's profile under poor visibility. However, the radiation temperature extracted possesses much less appearance features in comparison to visual cameras. State-of-the-art DeepTIO [34] incorporates thermal camera, IMU, and RGB visual hallucination to complement barren features from thermal alone.

We take the above deep odometry models as frontend candidates of our ubiquitous positioning system for their excellent capabilities in pedestrian positioning. These deep odometry models, with proven robustness in different adverse conditions, have drastically different computational complexity. Our next step is to deploy these deep odometry models on edge with a repeatable way.

### B. Edge Deployment

DNNs are computationally expensive. For instance, there are over 33 million parameters in milliEgo [29]. Its peak RAM usage exceeds $2GB$ running in Keras with TensorFlow backend. Careful forging of the inference engine is vital to success. The core of this is to elaborate a pipeline of sensor data fusion, synchronization, and inference engine which should be systematically optimized with a stable throughput. Meanwhile, deep odometry models based on different sensors could be developed in any Machine Learning framework to be embedded in a variety of edge devices. Common deployment methods are yet limited to a one-off network minimization upon a specific piece of hardware, such as knowledge distillation [7]. Model compression methods, e.g., quantization, compromise accuracy which hinders their usability in positioning systems. To facilitate a universal and lossless deployment process, it is imperative to establish a pipeline that maximizes interoperability for the deep odometry models.

We explore existing toolbox provided by Machine Learning frameworks to deploy the deep odometry models on edge. We note the combination of Open Neural Network Exchange (ONNX) [28] and TensorRT [8] makes an ideal model compression pipeline. ONNX is a universal Deep Learning framework that represent DNN in a universal format initiated by Microsoft and Facebook. We integrate ONNX as an intermediate layer in our deployment pipeline to transfer deep odometry models originally developed in any framework including Pytorch, TensorFlow, Keras etc. Since the corre-

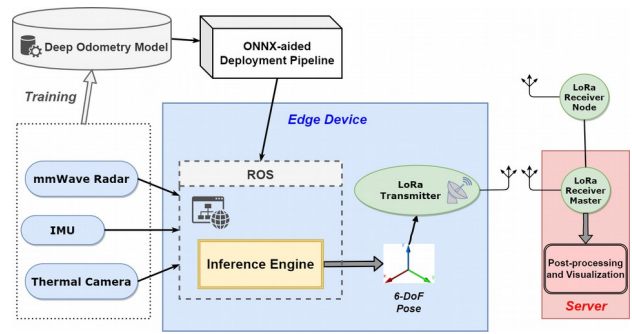

Fig. 3: A high-level architecture of the proposed positioning system. Sensor fusion and inference are performed on edge. A server conducts post-processing, data storage, and visualization.

TABLE I: Multi-Modalities Deep Odometry

A general comparison of deep odometry models that claim effective regardless of visual degradation based on their performance, power, and price.

|  | IONet (IMU) | MilliEgo (IMU + mmWave radar) | DeepTIO (IMU + thermal camera) |
|---|---|---|---|
| *Accuracy* | Less than $5m^\star$ | Less than $1.5m^\star$ | Less than $1m^\star$ |
| *Computational Complexity* | Very Low | Low | High |
| *Cost* | Low | Low | Relatively High |
| *Robustness and Reliability* | Agnostic of exterior environment; Subject to cumulative drift in the long term | Immune to illumination conditions; Resilient to motion noise; Require reflectors in close proximity | Resilient to glare or darkness; Sensitive to thermal features |

$\star$ throughout pedestrian walking tests within a few minutes.

sponding ONNX-Runtime is not efficient for inference, we use TensorRT, an NVIDIA proprietary framework, as the inference engine vendor. TensorRT applies graph optimizations, layer fusion, and finding the fastest implementation of a given DNN model. This allows the deep odometry frontend to make full use of the computational power on edge if GPU or Deep Learning Accelerator (DLA) is available. Flowchart of our proposed deployment pipeline is shown in Fig. 2.

Online sensor fusion and synchronization are accomplished via ROS, a *de-facto* robotic framework which allows simple system integration. A ROS-based architecture handles sensor and inference engine interconnection as a network of nodes. A high-level diagram of the system integration via ROS is displayed in Fig. 3.

### C. EKF-LoRa Backend

*1) LoRa:* There is need to send pedestrian's positional data back to the server. We utilize LoRa for its long range in NLOS transmission and a highly configurable bandwidth [14]. The RSSI that comes free with each LoRa packet provides an imperative link to project the deep odometry's aggregated positioning to the global frame. Thereby, it is of great importance to determine a path loss model to infer range from RSSI. We measured LoRa RSSIs as a function of distance using SF from 7 to 11 in an open area. We recorded over 1.1k LoRa packets per SF in various LOS and NLOS scenarios. As is seen in Fig. 4, SF 7 has a larger gradient, $\frac{\partial RSSI}{\partial d}$, implying a better range-indicative sensitivity compared to SF 11. We also measured their average air times in LOS condition when transmitting a 240-bytes message as shown in Table II. Though larger SFs tolerates low SNRs better, severe latency occurs.

Since a ubiquitous positioning system does not make a-*priori* assumption of the environment. We adopt a generic path loss model with attenuation factor, $n$. The path loss model as a function of transmitter-receiver distance, $d$, can be expressed as;

TABLE II: LoRa Air Time Benchmark

| SF | 7 | 8 | 9 | 10 | 11 |
|---|---|---|---|---|---|
| **Air Speed (Kpbs)** | 62.5 | 19.2 | 9.6 | 4.8 | 1.2 |
| **Air Time (s)** | 0.447 | 0.551 | 0.674 | 0.86 | 1.708 |

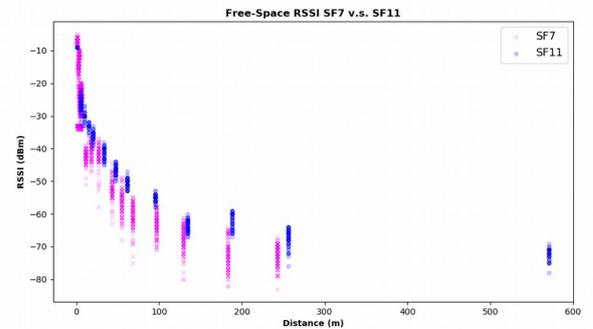

Fig. 4: LoRa RSSI open-area measurement at SF 7 and SF 11. The former shows a better range-indicative sensitivity.

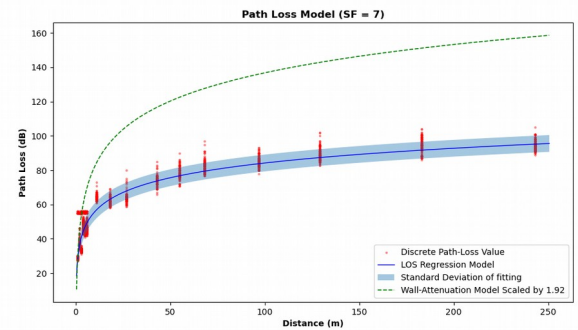

Fig. 5: LoRa path loss model at SF 7. Blue line shows the regression result based on measurement; Light blue area marks regression samples within the standard deviation; Dashed green line shows an empirically scaled path loss model for NLOS transmission.

$$PL(d) = PL(d_0) + 10n \cdot \log\frac{d}{d_0} + C \qquad (1)$$

where $d_0$ is the reference distance, $PL(d_0)$ is assumed the free space path loss at the reference distance in dB, and $C$ is a constant to calibrate bias. We take $d_0 = 1m$ as reference and obtain a mean RSSI of $-8.483dBm$. Given a transmitting power of $22dBm$, it is determined $PL(d_0) = 30.483dB$. The attenuation factor and bias are derived through polynomial regression as shown in Fig. 5, in which the standard deviation equals $4.887dB$. To extract range from measurement, the RSSI can be directly expressed as a function of distance as below;

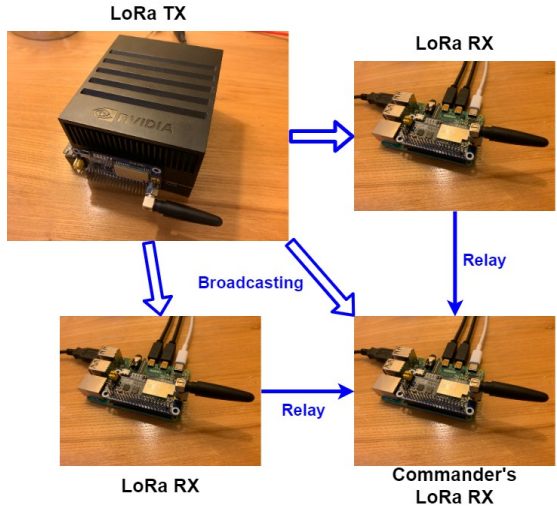

Fig. 6: LoRa network topology. The transmitter is connected to the edge device, e.g., an NVIDIA Jetson AGX Xavier. One LoRa node/gateway is designated as server at the mission commander's end. Other LoRa nodes are expected to relay LoRa RSSIs to the server.

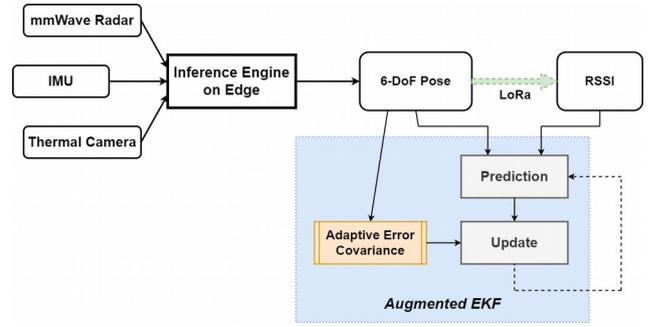

Fig. 7: An overview of online data flow and integration of the augmented EKF.

$$RSSI(d) = -28.5737 \times \log_{10}(d) - 5.06 \quad (2)$$

It is assumed the RSSI samples are log-normally distributed. Thus, the probability distributed of range is regarded Gaussian;

$$RSSI\ [dBm] \sim N(\mu_{rssi}, \hat{\delta}) \quad (3)$$

where $\delta$ is the standard deviation of measurement, i.e., equivalent to the regression error, i.e., $4.887dB$. It can also be used as generic uncertainty of any RSSI observation from the field [35].

Since our proposed positioning system aims at adverse indoor or post-disaster environment, we presume NLOS as the predominant use cases in which a free-space path loss model tends to underestimate. W. Xu *et al.* [41] propose experimental path loss scaling factors when LoRa transmits through walls of different materials. We take an empirical attenuation scaling factor, $\alpha = 1.92$, on the basis of our path loss model. The final LoRa path loss model is marked as dashed green line in Fig. 5.

One LoRa node or gateway is designated as the server where the real-time positioning data shall be informed, stored, and visualized as illustrated in Fig. 3. Note the LoRa transmitter constantly broadcasts packets to the server and other anchor nodes. Since the broadcast packet should be identical, only the RSSIs from peripheral anchors are relayed to the server. The topology of this setup is illustrated in Fig. 6.

*2) EKF:* We argue that a generic EKF backend is sufficient to fuse the pose of any deep odometry candidate and the RSSIs as range constraints, so as to a refined trajectory. EKF is not only robust in handling non-linear motion modelling [15] but extremely lightweight, thus, suitable for online systems. A rigid transformation $\mathbf{T}_{k,k-1} \in \mathbb{R}^{4\times4}$ of two consecutive deep odometry poses from time $k-1$ to $k$ can be expressed as;

$$\mathbf{T}_{k,k-1} = \begin{bmatrix} \mathbf{R}_{k,k-1} & \mathbf{t}_{k,k-1} \\ 0 & 1 \end{bmatrix} \quad (4)$$

where $\mathbf{R}_{k,k-1}$ is the rotation matrix at time $k$, and $\mathbf{t}_{k,k-1}$ is the translation matrix. Given a set of poses from the starting position $\mathbf{C}_{0:k} = \{\mathbf{C}_0, \mathbf{C}_1, \ldots, \mathbf{C}_k\}$, the current pose at time $k$ can be written as;

$$\mathbf{C}_k = \mathbf{C}_{k-1}\mathbf{T}_{k,k-1} \quad (5)$$

Hence, the role of deep odometry is to calculate $\mathbf{T}_k$ according to the inertial and exteroceptive sensors in order to restore the path $\mathbf{C}_{0:k}$ [20].

To this end, the goal of the EKF-LoRa backend is to minimize the disparities, $\varepsilon$, of ranges from $M$ anchors and the deep odometry pose by deriving a posterior $\hat{\mathbf{C}}$. For time $k$, this can be written as;

$$\varepsilon_k = \underset{\hat{\mathbf{C}}_k}{\operatorname{argmin}} \sum_{m=1}^{M} |\|\mathbf{C}_k - \mathbf{p}_m\| - d_m| \quad (6)$$

where $\mathbf{p}_m$ is the pose of the $m$-th anchor, and $d_m$ is the distance from the $m$-th anchor denoted in Eqn. 1.

In this work, we consider the positioning fusion in 2D domain for simplicity. Note it can be easily adapted to 3D localization. We specify the current pose as system state and RSSIs as observation of the EKF. The system state shall be updated by new observations. The schematic diagram is shown in Fig. 7. The system state is made up of 2D coordinates and azimuth heading direction, i.e., $(x, y)$ and $\theta$. It reads

$$\mathbf{x} = [x, y, \bar{\theta}] \quad (7)$$

The state transition and measurement equations can be expressed as

$$\mathbf{x}_k = f(\mathbf{x}_{k-1}) + \omega_{k-1} \quad (8)$$
$$\mathbf{z}_k = h(\mathbf{x}_k) + \nu_k \quad (9)$$

where $\omega_{k-1}$ is the process noise with covariance $\mathbf{Q}$, $\mathbf{z}_k$ is the measurement vector, and $\nu_k$ is the measurement noise. Assume the process noise is of zero-mean Gaussian, we specify the initial state variance as $\mathbf{Q} = \mathbf{diag}\,[0.1, 0.1, 0.01]$.

We represent the 6-DoF deep odometry pose as a 4-by-4 affine transformation matrix, $\mathbf{C_k}$, at time $k$. The initial pose is defined as

$$\mathbf{C_0} = \begin{bmatrix} 1 & 0 & 0 & x_0 \\ 0 & 1 & 0 & y_0 \\ 0 & 0 & 1 & 0 \\ 0 & 0 & 0 & 1 \end{bmatrix} \quad (10)$$

where $(x_0, y_0)$ marks the starting position with initial orientation towards positive $x$-axis. The aggregate pose follows Eqn. 5.

As a result, the prior state which can be extracted from $\mathbf{C_k}$ contains relative 2D position, current yaw angle, and the RSSIs from $M$ anchors, denoted as $\xi_x$, $\xi_y$, $\xi_\theta$, and $\gamma_i$, respectively;

$$\mathbf{z_k} = [\xi_x, \xi_y, \xi_\theta, \gamma_1, \gamma_2, \ldots \gamma_M]^T \quad (11)$$

The state transition model can be written as an identity matrix plus process noise;

$$\mathbf{x_k} = f(\mathbf{x_{k-1}}) \approx \mathbf{I}_{3 \times 3} \cdot \mathbf{x_{k-1}} + \omega_{k-1} \quad (12)$$

The predicted error covariance matrix yields

$$\hat{\mathbf{P}}_\mathbf{k} = \mathbf{F_{k-1}} \check{\mathbf{P}}_\mathbf{k-1} \mathbf{F_{k-1}}^T + \mathbf{Q} \quad (13)$$

where a large initial error covariance, $\mathbf{P_0} = \mathbf{diag}\,[10, 10, 1]$ is assigned to allow convergence from an unknown starting position. In the update stage, the Kalman Gain is defined as

$$\mathbf{K_k} = \check{\mathbf{P}}_\mathbf{k} \mathbf{H_k}^T \left[ \mathbf{H_k} \check{\mathbf{P}}_\mathbf{k} \mathbf{H_k}^T + \mathbf{R} \right]^{-1} \quad (14)$$

where $\mathbf{H_k}$ is the Jacobian matrix of the measurement in Eqn. 9, and $\mathbf{R}$ is error covariance of the measurement noise equal to

$$\mathbf{R} = \mathbf{diag}\,[\sigma_x^2, \sigma_y^2, \sigma_\theta^2, \delta^2 \ldots] \quad (15)$$

where $\delta = 4.887$ holds for all RSSI measurements derived from path loss regression.

We note that deep odometry errors are correlated with a high body turning speed [11]. Since $\sigma_x$ and $\sigma_y$ are associated with translation only, we assign small error covariance, $\sigma_x = \sigma_y = 0.15$ to minimize error from experiments. On the other hand, the error covariance of heading is directly related to turning. We augment the EKF by associating the error covariance of heading, $\sigma_\theta^2$, to current angular velocity at time $k$;

$$\omega_k = \frac{d\mathbf{R}_{k,k-1}}{d\triangle t} \quad (16)$$

where $\triangle t$ is the frame interval. Therefore, we make the error covariance of the heading as a function of $\omega_k$;

$$\sigma_\theta^2 = 10 \cdot \left(1 - \bar{e}^{0.2 \cdot \|\omega_k\|}\right) \quad (17)$$

in doing so, we realize stable accuracy gain in real-time tests with various turning speeds.

Since RSSIs are noisy, we apply a weighted average filter to smooth the RSSIs so as to stabilize range estimates. Given a series of RSSIs observed at time $k$, $\gamma$, the filter can be written as

$$\overline{\gamma_k} = \sum_{i=0}^{\lambda} \gamma_{k-i} \cdot \frac{1}{\lambda} \quad (18)$$

where the window size $\lambda = 4$. An RSSI can be expressed as a function of range according to Eqn. 2;

$$\gamma_i = \alpha \cdot \log\sqrt{(x - x_i)^2 + (y - y_i)^2 + (0 - z_i)^2} + \beta \quad (19)$$

where $\alpha = -28.5737$, $\beta = -5.06$, and $\mathbf{p_i} = [x_i, y_i, z_i]$ are the $i$-th known anchor coordinates. Denote $\mathbf{p} = [x, y, 0]$ as current position, the Jacobian matrix can be written as

$$\mathbf{H} = \left.\frac{\partial h}{\partial \mathbf{x}}\right|_{\hat{\mathbf{x}}_k} = \begin{bmatrix} 1 & 0 & 0 \\ 0 & 1 & 0 \\ 0 & 0 & 1 \\ \frac{\alpha(x-x_i)}{\ln 10 \|\mathbf{p}-\mathbf{p_i}\|^2} & \frac{\alpha(y-y_i)}{\ln 10 \|\mathbf{p}-\mathbf{p_i}\|^2} & 0 \\ \ldots & \ldots & \end{bmatrix} \quad (20)$$

Hence, the measurement residual can be used to update the state

$$\hat{\mathbf{x}}_\mathbf{k} = \check{\mathbf{x}}_\mathbf{k} + \mathbf{K_k} \cdot (\mathbf{z_k} - h(\check{\mathbf{x}}_\mathbf{k})) \quad (21)$$

and the updated error covariance matrix can be written as

$$\hat{\mathbf{P}}_\mathbf{k} = (\mathbf{I} - \mathbf{K_k} \cdot \mathbf{H_k}) \check{\mathbf{P}}_\mathbf{k} \quad (22)$$

Pedestrians may start being tracked from anywhere in the reference/global coordinate system. The initial guess of the system state, $\mathbf{x_0}$, may take arbitrary values in which it would converge towards the true state given more measurements. We conduct extensive tests in various environments to validate our proposed positioning system in the next section.

## IV. EVALUATION

For a fair comparison, we collected 8 hours of synchronized thermal, mmWave radar, and IMU data to train the deep odometry models, in which a 20% split is used for validation. We labelled the multi-sensor data using reconstructed trajectories of a Velodyne HDL-32E LiDAR. We trained these models, namely IONet, milliEgo, and DeepTIO, at 10 FPS for 200 epochs to get the best convergence. We setup six SX1262 LoRa modules stacked on Raspberry Pi 4s as transceivers. Whenever receiving a packet, a LoRa receiver node predicts an RSSI, in $dBm$ with $1dB$ resolution, with no additional equipment required.

We carried out a series of tests in unsurveyed environments to validate the proposed positioning system. We evaluated these deep odometry candidates across a building floor with corridors and a poorly illuminated room, and a ground-floor apartment with access to an outdoor car park. The performance attributes investigated include accuracy, frame rate, number of LoRa anchors used, range of effectiveness, and system latency.

## A. Accuracy and Robustness

To evaluate the performance in adverse conditions, three experiment scenarios are chosen including a square-shape corridor, an office with no light on, and an outdoor car park connected to a ground-floor apartment. The former two were carried out in the same building floor with a size of $20m \times 30m$. In the 3rd scenario, the experimenter walked in and out of the ground-floor apartment to an open car park at night. Four to five LoRa anchors were arranged arbitrarily nearby the walking area within $50m$. The experimenter kept walking for 2 to 3 minutes during each test.

The Root-Mean-Square Errors (RMSE) of the aforementioned experiments are shown in Table III, Table IV, and Table V, respectively. Without EKF-LoRa backend, i.e., 'Anchor = 0', IONet incurs large drifting errors; MilliEgo and DeepTIO remain resilient in all three testing scenarios. It indicates that standalone milliEgo produce decent accuracy comparable to DeepTIO, given milliEgo is much more compact.

An EKF-LoRa calibrated trajectory compared to using milliEgo alone in the office test is plotted in Fig. 8. The calibrated and original trajectories using DeepTIO in apartment and car park test are plotted in Fig. 9. Notice the accuracy sees decent increment with even one anchor which verifies our hypothesis.

As more RSSIs from different LoRa anchors are taken into positioning fusion, better accuracy is seen for all deep odometry candidates throughout all tests. With at least three LoRa anchors, the positioning errors after calibration were almost halved. Specifically, given three anchors IONet, milliEgo, and DeepTIO see an average of $69\%$ $34\%$ and $36\%$ accuracy gains, respectively. This demonstrates the efficacy of our proposed positioning system in different environments based upon any deep odometry frontend. When there are more than three anchors, the benefit of having more LoRa anchors becomes marginal.

## B. Odometry Selection and Frame Rate

Lightweight deep odometry systems are more desirable for real-time applications on edge. Similarly, a low frame rate linearly reduces the computational burden. We investigate the effect of halving the original training and inference frame rate of the deep odometry candidates. The results are shown in Table VI. We repeat the corridor and car park tests using skipped sensor data stream recorded. The trajectory of IONet at 5 FPS in the corridor scenario is shown in Fig. 10. It can be seen the accuracy of either deep odometry model deteriorates at 5 FPS. IONet finds a significant deviation from ground-truth because it requires a high refresh rate to capture human motion. Nevertheless, milliEgo at 5 FPS pertains a decent tracking accuracy which lends itself a power-efficient choice under extreme resource constraints, even running on a Raspberry Pi. Overall, we believe milliEgo, i.e., mmWave radar plus IMU, makes an optimal positioning solution for adverse environment in terms of accuracy and efficiency. Note we omit the frame rate test on DeepTIO for training it at 5 FPS does not converge well.

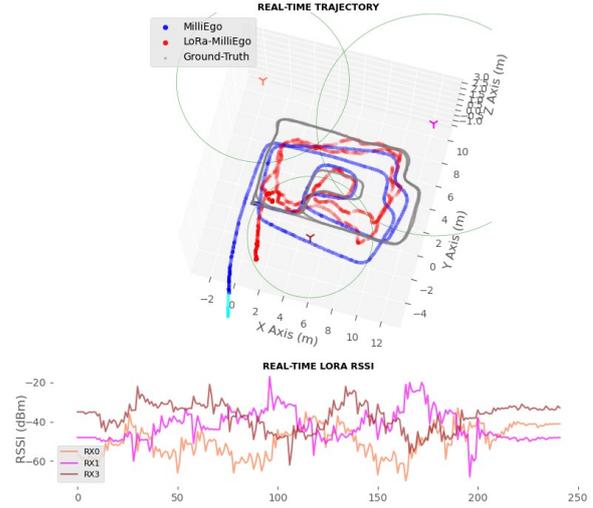

Fig. 8: Trajectory of a two-round search along corridors and entering and exiting a dark room. The ground-truth path is marked in grey; Blue path indicates using milliEgo only; Red path indicates fusion of milliEgo and LoRa RSSIs; The light green circles specify range estimates derived from the path loss model with three LoRa anchors painted in different colours.

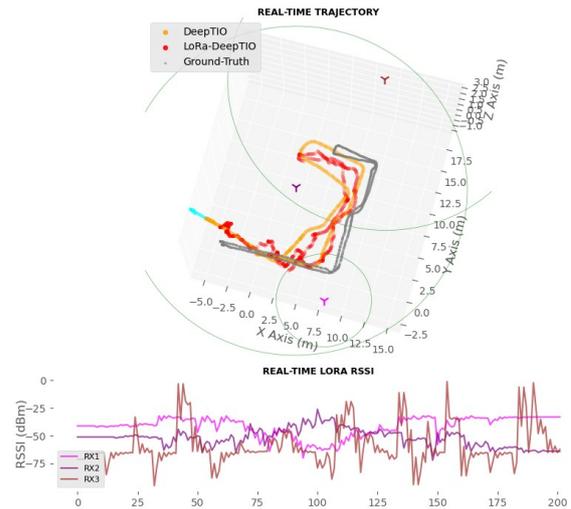

Fig. 9: Trajectory of a search in a ground-floor apartment and an outdoor car park; The ground-truth path is marked in grey; Brown path indicates using DeepTIO only; Red path indicates fusion of DeepTIO and LoRa RSSIs; The light green circles specify range estimates derived from the path loss model with three LoRa anchors painted in different colours.

TABLE III: RMSE (m) among IONet, MilliEgo, and DeepTIO on three-round search along a square-shaped corridor

| RMSE (m) | Anchor = 0 | Anchor = 1 | | Anchor = 3 | | | | Anchor = 4 | | Anchor = 5 |
|---|---|---|---|---|---|---|---|---|---|---|
| | | Test i | Test ii | Test i | Test ii | Test iii | Test iv | Test i | Test ii | |
| IONet | 17.603 | 12.87 | 13.975 | 6.741 | 5.679 | 9.559 | 9.04 | 5.731 | 6.024 | **4.03** |
| MilliEgo | 3.202 | 2.628 | 3.033 | 2.067 | 2.437 | 2.15 | 3.768 | 2.687 | 2.404 | **1.959** |
| DeepTIO | 3.517 | 3.461 | 2.977 | 2.024 | 2.858 | 1.95 | 2.146 | 1.713 | 2.052 | **1.612** |

TABLE IV: RMSE (m) among IONet, MilliEgo, and DeepTIO on entering and exiting a dark office

| RMSE (m) | Anchor = 0 | Anchor = 1 | | Anchor = 3 | | | | Anchor = 4 |
|---|---|---|---|---|---|---|---|---|
| | | Test i | Test ii | Test Test i | Test ii | Test iii | Test iv | |
| IONet | 15.237 | 12.674 | 12.835 | 8.979 | 9.956 | 8.33 | 10.031 | **6.231** |
| MilliEgo | 3.574 | 2.798 | 2.634 | 2.394 | 3.203 | 3.142 | **2.312** | 2.617 |
| DeepTIO | 3.122 | 3.223 | 2.841 | **2.034** | 2.676 | 3.055 | 2.971 | 2.629 |

TABLE V: RMSE (m) among IONet, MilliEgo, and DeepTIO on wandering in a ground-floor apartment and a car park at night

| RMSE (m) | Anchor = 0 | Anchor = 1 | | Anchor = 3 | | | | Anchor = 4 |
|---|---|---|---|---|---|---|---|---|
| | | Test i | Test ii | Test Test i | Test ii | Test iii | Test iv | |
| IONet | 20.609 | 15.882 | 13.09 | 13.214 | 16.348 | 14.653 | 14.484 | **12.477** |
| MilliEgo | 3.789 | 3.012 | 3.116 | 3.385 | 2.849 | **2.826** | 3.48 | 3.063 |
| DeepTIO | 3.682 | 3.598 | 3.746 | 2.411 | 2.586 | 2.71 | 2.542 | **2.346** |

TABLE VI: Effect of Frame Rate to Deep Odometry

| RMSE (m) | | Corridor | Apartment and Cap Park |
|---|---|---|---|
| IONet | 5 FPS | 21.04 | 35.464 |
| | 10 FPS | 17.603 | 20.609 |
| milliEgo | 5 FPS | 6.277 | 4.483 |
| | 10 FPS | 3.202 | 3.789 |

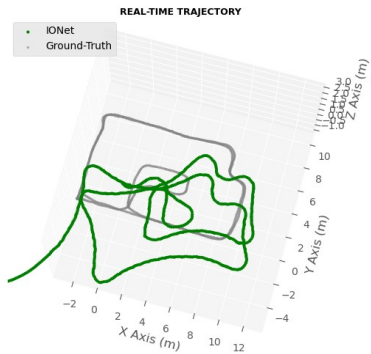

Fig. 10: Trajectory of using IONet at 5 FPS in the corridor test in comparison to the ground-truth. The results were generated using the same sensor data stream recorded by Rosbag. Though a faithful shape of trajectory is pertained, IONet suffers from larger drifting errors.

### C. System Latency and Transmission Range

A Raspberry Pi 4 can be an extremely compact and low-cost platform despite a CPU-only platform. In contrary, the NVIDIA AGX Xavier has on-board GPU and DLA with the capability of a desktop enclosed in an embedded module under $30W$. We take the selection of device into consideration and investigate runtime performance of three representative edge devices, namely Raspberry Pi 4, NIVDIA Jetson TX2, and NVIDIA Jetson AGX Xavier. The latency benchmark results are shown in Table VII.

As can be seen, the Jetson AGX Xavier delivers the highest computational efficiency with TensorRT. It makes 10 FPS achievable for deep odometry models as complex as DeepTIO. The Jetson AGX Xavier is also the only platform currently supporting latest version of TensorRT. An exceptionally low latency of $3.5ms$ is seen running milliEgo. The inference engine loading time is also found reducing to $4.4s$ as opposed to approximately $40s$ using Keras with TensorFlow backend. The Jetson TX2, with fewer GPU cores and no DLA on board, produces relatively low delays. With the poorest compatibility, Raspberry Pi 4 finds the highest latency given it is a CPU-only platform. Nevertheless, it is possible to run IONet at 5 FPS on Raspberry Pi 4.

Though configuring LoRa at SF 7 grants a rather low air time and a higher bandwidth, the potential transmission range is compromised, especially when node-to-node communication is used in stead of having a gateway. To test the coverage of LoRa node-to-node transmission, we conducted an experiment by walking away from an SX1262 LoRa transmitter sitting indoor, broadcasting at $22dBm$. We walked away along four directions in the built-up area and observed no packet loss within $260m$ in any direction. This is sufficient for contingent positioning service on site. Apparently, using a LoRa gateway with active antennas or configuring a high SF allows much greater range.

As for the server, the processing of the EKF and visualization (based on *PyQt5*) of a pedestrian's location take approximately $210ms$ on a Dell XPS 15 laptop with an Intel i7 processor. Given a Jetson AGX Xavier as edge platform and LoRa SF 7, it takes less than $1s$ to visualize the pedestrian's location. This proves our proposed positioning system is suitable for real-time applications.

## V. DISCUSSION

The augmented EKF is a tightly-coupled method to fuse deep odometry and LoRa RSSIs. This assumes all observations are legitimate with a degree of Gaussian noise. Yet outliers occur when the transmission path finds unexpected interference or sensor data are corrupted. In the next step, we will implement an outlier detection layer to decouple the EKF from updating to further improve the system stability.

TABLE VII: Inference latency benchmark

|  | **Mean Latency (ms)** | NIVDIA Jetson AGX Xavier | NVIDIA Jetson TX2 | Raspberry Pi 4 |
|---|---|---|---|---|
| IONet | Keras with TensorFlow backend | 21.9 | 75.3 | 110 |
| MilliEgo | Keras with TensorFlow backend | 25.8 | 57.9 | 379 |
|  | ONNX-Runtime | 10.4 | 141 | N/A |
|  | TensorRT serialized engine | 3.5 | N/A | N/A |
| DeepTIO | Keras with TensorFlow backend | 71.9 | 237 | 10676 |
|  | TensorRT serialized engine | 21.1 | N/A | N/A |

If LoRa anchors are manually placed in a horizontal plane, the range-calibrating effect of LoRa in 3D positioning is significantly restrained. Furthermore, it is most likely the anchors end up in sub-optimal locations. To overcome these drawbacks, we plan to devise UAVs as mobile anchors for 3D positioning. Optimal anchor locations may be calculated at runtime to allow minimize the pedestrian's positioning uncertainty. Meanwhile, other wireless metrics, such as TDOA and AOA, can be used to further enhance positioning accuracy.

## VI. CONCLUSION

This paper presents a novel edge-deployable ubiquitous positioning system based on deep odometry and an augmented EKF-LoRa backend. We reveal the feasibility of utilizing these deep odometry to infer position in real-time under various resource constraints. We design a universal pipeline of deploying arbitrary deep odometry model on edge. To bring positional data from odometry that aggregate in local frame to the global frame and mitigate the drifting errors, an EKF-LoRa fusion module is proposed. We find a consistent accuracy gain, over 34% across all tested deep odometry models. Considering the experimental results regarding accuracy and efficiency, we believe the combination of mmWave rader and IMU make the best solution for pedestrian localization in adverse environment.